\documentclass{article}

\usepackage[preprint]{neurips_2020}
\usepackage{xcolor, soul, mathtools, tabularx, multirow, amsmath}
\DeclareMathOperator{\diag}{diag}

\usepackage[utf8]{inputenc} % allow utf-8 input
\usepackage[T1]{fontenc}    % use 8-bit T1 fonts
\usepackage{hyperref}       % hyperlinks
\usepackage{url}            % simple URL typesetting
\usepackage{booktabs}       % professional-quality tables
\usepackage{amsfonts}       % blackboard math symbols
\usepackage{nicefrac}       % compact symbols for 1/2, etc.
\usepackage{microtype}      % microtypography

\title{Label-GCN: An Effective Method for Adding Label Propagation to Graph Convolutional Networks}

\author{
  Claudio Bellei\\
  Elliptic\\
  London, UK \\
  \texttt{claudio@elliptic.co} \\
  \And
  Hussain Alattas \\
  Department of Computer Science  \\
  The University of Sheffield, UK \\
  \And
  Nesrine Kaaniche \\
  Telecom SudParis \\
  Polytechnic Institute of Paris
}

\begin{document}

\maketitle

\begin{abstract}
We show that a modification of the first layer of a Graph Convolutional Network (GCN) can be used to effectively propagate label information across neighbor nodes, for binary and multi-class classification problems. This is done by selectively eliminating self-loops for the label features during the training phase of a GCN. The GCN architecture is otherwise unchanged, without any extra hyper-parameters, and can be used in both a transductive and inductive setting. We show through several experiments that, depending on how many labels are available during the inference phase, this strategy can lead to a substantial improvement in the model performance compared to a standard GCN approach, including with imbalanced datasets. 
\end{abstract}

\section{Introduction}
In recent years, the emergence and development of graph neural networks (GNNs) has allowed the extension of deep learning methods to data that have an underlying non-Eucledian structure. Many variants of GNNs have been proposed over the years, with applications ranging from recommender systems, to chemistry, traffic control, physics, and more \cite{Zhou2018}.
One of such variants, Graph Convolutional Networks (GCN) \cite{Kipf2017}, can be considered a state-of-the-art model for learning graph representations \cite{Shchur2019}. GCNs are multi-layer feedforward neural networks where, at each layer, a first-order approximation of a spectral graph convolution is applied. GCNs can be applied to semi-supervised node classification problems, where the graph is made of both labeled and unlabeled data and the goal is to predict the label of a node, given the graph structure and a set of features associated with each node. 

In GCN, labels are not used in the learning phase and therefore this information does not affect the prediction of an unknown node. However, in many applications, nearby nodes or nodes that belong to the same graph structure might share the same label. For such applications, one would expect that the ability to use label information should lead to an improvement of the model.

In this paper, we show that a simple modification of the first layer of a GCN can make the model learn information from neighbor labels. This leads to a measurable improvement in performance in various experiments of citation and Bitcoin networks, depending on the availability of neighbor labels during the inference phase. The model maintains the properties of the original GCN approach, i.e., it does not add extra hyper-parameters and can be used in both a transductive setting - where the whole graph structure is known during the training phase - and an inductive setting - where the model is tested on previously unseen nodes. Supporting code is available at \href{https://github.com/cbellei/LabelGCN}{https://github.com/cbellei/LabelGCN}.

\section{Related work}

The notion that nearby nodes of a graph are likely to have the same label has been modeled by researchers over the years through various variants of an algorithm that goes under the name of Label Propagation (LP) \cite{Zhu2002, Szummer2002, Zhou2003, Wang2009, Wang2013, Wu2012, Karasuyama2013, Gong2017, Viswanathan2019}. LP is a semi-supervised learning algorithm that propagates known labels along the edges of a graph, for the purpose of classifying unlabelled nodes. Examples of real-world datasets where information from neighbor labels has been shown to be beneficial include social networks, web pages, protein-protein interactions, citation graphs and anti-money laundering in the Bitcoin network \cite{Du2013, Viswanathan2019,  Raghavan2007, Hu2019}. 

The idea of combining graph convolutions with label propagation has only been explored recently \cite{Wang2020, Dong2020, Ragesh2020}. In \cite{Wang2020}, a model is proposed that unifies LP and GCN. There is a duality between the two models, in that LP can be interpreted as propagating \textit{labels} along edges in the graph and averaging them linearly, while GCN as propagating \textit{features} along the edges and combining them non-linearly through the various layers that compose the neural network. The original model proposed in \cite{Wang2020} works in a transductive setting, where all node features and graph structure is available during the training phase; the GCN model has to learn both the weight matrix and a weighted adjacency matrix, which can lead to overfitting. In \cite{Dong2020}, a two-step process is proposed, that includes a first step of label propagation that feeds into a second step of a neural network classifier. Similarly in \cite{Ragesh2020}, a general framework for composing networks is discussed, which includes the possibility of composing GCN with a label propagation network. 

The present paper proposes an alternative method to adding label propagation to GCN, through a simple variant of the original GCN approach. This has three main advantages: 1. it allows the model to work in an inductive setting, where the testing nodes are not part of the graph structure during the training phase; 2. it does not result in the addition of new hyper-parameters, which often increases the risk of overfitting the training data; 3. it does not add any significant algorithmic complexity to the original GCN approach. 

\section{Label-GCN}
\subsection{Graph Convolutional Networks}
\label{sec:overviewgcn}

Graph Convolutional Networks \cite{Kipf2017} are a class of multi-layer neural network architectures designed to take advantage of the graph structure of a given dataset. The graph structure is specified by using an undirected adjacency matrix $A\in\mathbb{R}^{n\times n}$, where $n$ is the number of nodes in the graph, along with the input matrix $X\in\mathbb{R}^{n\times d}$ containing the features for each node, where $d$ is the dimension of the feature vector. The output of every new hidden layer in GCN combines a normalized adjacency matrix $\widehat{A}$, the output of the previous hidden layer $H^{(l)} \in \mathbb{R}^{n\times d_l}$ and a weight matrix $W^{(l)}\in \mathbb{R}^{d_l \times d_{l+1}}$,

\begin{equation}
H^{(l+1)}=\sigma\left(\widehat{A} H^{(l)} W^{(l)}\right)
\label{e14}
\end{equation}

\begin{equation}
\widehat{A}=\widetilde{D}^{-1/2} \widetilde{A}\,\widetilde{D}^{-1/2}, \quad \widetilde{A}=A+I, \quad \widetilde{D}_{ii}=\sum_{j} \widetilde{A}_{ij}
\label{e15}
\end{equation}
The first hidden layer corresponds to the input matrix, $H^{(0)} \coloneqq X$. The $\sigma$ operator in eq.~(\ref{e14}) represents a non-linear activation function (typically ReLU). The normalized adjacency matrix $\widehat{A}$ differs from the standard adjacency matrix $A$ in that self-loops are added and the matrix elements are rescaled through the diagonal node degree matrix $\widetilde{D}$.   
\par Over time, many variants of the original GCN approach have been proposed. Notable ones have been directed at improving the applicability of graph-based models for large scale applications \cite{Hamilton2017, Chen2018, Wu2019}. A recent survey on the state of the research can be found in \cite{Zhang2020}.

\subsection{Label-GCN}
\label{sec:modifgcn}
For many applications, it is reasonable to assume that knowledge of the labels surrounding an unknown center node can help predict the label of the center node itself. This suggests adding a new feature during learning of the network: the label of each node. However in the canonical formulation of GCN, eqs.~(\ref{e14}) and (\ref{e15}), the convolution operation would include a self-loop for each center node. Simply adding the label as a new "feature" during the training phase of the model would clearly not work: the model would simply learn to use the label for the prediction, which would not be available during the testing phase. What is instead desirable is to train the network through the neighbors' labels, without addition of the self-loops. 
\par In order to achieve this, it is possible to modify the first layer of a GCN to include the labels of the neighbors, for each center node, but not the label of the center node itself. Labels can then become part of the feature vector, as long as the model is only allowed to see the labels of the neighbors. For a classification problem with $K$ classes, assuming one-hot encoding the dimensionality of the input matrix therefore increases by a factor $K$, such that $X\in\mathbb{R}^{n\times (d + K)}$. The modified GCN network that achieves the required behavior is then
\begin{align}
& H^{(1)} = \sigma\left[\left(\widehat{A}X - \diag(\widehat{A})X\sum_{j=1}^K e_{j}e_{j}^T\right)W^{(0)}\right] \label{e_gcnmodified1} \\
& H^{(l+1)} = \sigma\left(\widehat{A}H^{(l)}W^{(l)}\right) \ \ l \geq 1 \label{e_gcnmodified2}
\end{align}
In eq.~(\ref{e_gcnmodified1}), $e_j$ is a unit vector that has all its components equal to zero, except for the index associated with the position of the one-hot encoded class within the feature vector, $e_{j} = [0, \dots, 0, 1, 0, \dots 0]^T$. The product $e_{j}e_{j}^T$ is a single-entry matrix $J^{jj}$ and the term $$-\diag(\widehat{A})X\sum_{j=1}^K e_{j}e_{j}^T$$ selectively eliminates the self-loops for the components of the feature vector corresponding to the labels. For all the other components, the standard convolution operation is applied.
\par
\begin{figure}[h]
  \centering
  \includegraphics[width=0.6\linewidth]{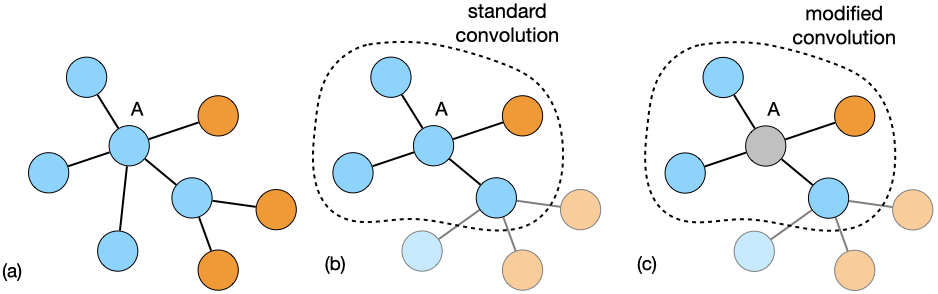}
  \caption{The difference between standard convolution, eq.~(\ref{e_gcnmodified2}), and the modified convolution, eq.~(\ref{e_gcnmodified1}), when self-loops are removed.}
  \label{fig:subgraph}
\end{figure}

 Training the network using eq.~(\ref{e_gcnmodified1}) corresponds to masking the label of a center node. For a visual explanation, consider node A in Figure \ref{fig:subgraph}a. The standard convolution, eq.~(\ref{e_gcnmodified2}), aggregates all the values around the center node, including the value of the center node itself (Figure \ref{fig:subgraph}b). The modified convolution of eq.~(\ref{e_gcnmodified1}) computes the same aggregation, except for the components of the feature vector corresponding to the labels: in this case it only aggregates the neighbors, avoiding the self-loop (Figure \ref{fig:subgraph}c).

\section{Experiments}
\label{sec:exp}

\subsection{Datasets}
For our experiments, we test our model on four different datasets, treating the graphs as undirected. The datasets are the well-known citation graphs (CORA, CiteSeer \cite{Sen2008} and PubMed \cite{Namata2012}) and the Elliptic dataset, a Bitcoin transaction graph \cite{Weber2019}. The statistics on the datasets are shown in Table \ref{tab:1}. Given that the Bitcoin network graph is more recent and less studied, we give below more details about it and motivate why label propagation can work for such problems. 

\begin{table*}[t]
  \centering  
  \resizebox{\linewidth}{!}{
  \begin{tabular}{lccccccc}
     \textbf{Dataset} &  \textbf{Type} & \textbf{Nodes} &  \textbf{Edges} & \textbf{Classes} & \textbf{Features} & \textbf{Training/Validation/Test/Support}\\ 
     \midrule
    \textbf{Cora} & Citation network & 2,708 & 5,429 & 7 & 1,433 & 140/140/273/2,155\\
    \textbf{Citeseer} & Citation network & 3,312 & 4,715 & 6 & 3,703 & 120/120/332/2,740\\
    \textbf{PubMed} & Citation network & 19,717 & 44,338 & 3 & 500 & 60/60/1973/17,624\\
    \textbf{Elliptic} & Bitcoin network & 203,769 & 234,355 & 2 & 166 & 4,656/4,656/9,314/27,938
  \end{tabular}
  }
  \caption{Dataset statistics and splits for the experiments (transductive setting). The concept of "support" set is explained in Subsection \ref{sec:setup}.}
  \label{tab:1}
\end{table*}

\subsubsection{The Elliptic Dataset}
The Elliptic dataset is a publicly available dataset that is made of 203,769 nodes and 234,355 edges, divided into 49 distinct time steps \cite{Elliptic2019}. Each node represents a transaction; an edge represents the flow of bitcoins between two transactions, as in Figure \ref{fig:propagation}b. There are 166 features associated with each node. The classes in the dataset are grouped by entities belonging to the "licit", "illicit" or "unknown" categories; the proportion of nodes in the dataset are 21\%, 2\% and 77\%, respectively. Therefore, this is an imbalanced dataset and only a fraction of it (23\% or 46,564 nodes) is labelled.

\subsubsection{Label propagation in the context of Bitcoin}
\label{sec:LAP}
The use of labels for the purpose of detecting illicit transactions in the Bitcoin network has been investigated in \cite{Hu2019, Oliveira2021}. In particular, in \cite{Oliveira2021} it is shown that this information can lead to a significant improvement of the results on the Elliptic dataset. Here, we want to motivate the reason why this is the case.

One of the main features of Bitcoin (as well as of other blockchains such as Zcash, Litecoin, Bitcoin Cash, Bitcoin SV, ...) is that it is unspent-transaction-output (UTXO) based: each input of a given transaction unlocks the funds associated with an output from a previous transaction. The transaction generates new (unspent) outputs that can be spent by the private-key holders of those outputs in future transactions. Typically, among those private-key holders, one of them corresponds to the same entity that has generated the transaction itself. The reason is that 1. normally the inputs of a transaction are controlled by the same entity and 2. what is left of a transaction goes back as one of the outputs to the same entity generating it (the "change" address of the transaction). These two observations have implications with respect to the privacy aspects of the Bitcoin blockchain \cite{Reid2013, Androulaki2013, Ron2013, Meiklejohn2016, Harrigan2016}.
\begin{figure}[h]
  \centering
  \includegraphics[width=\linewidth]{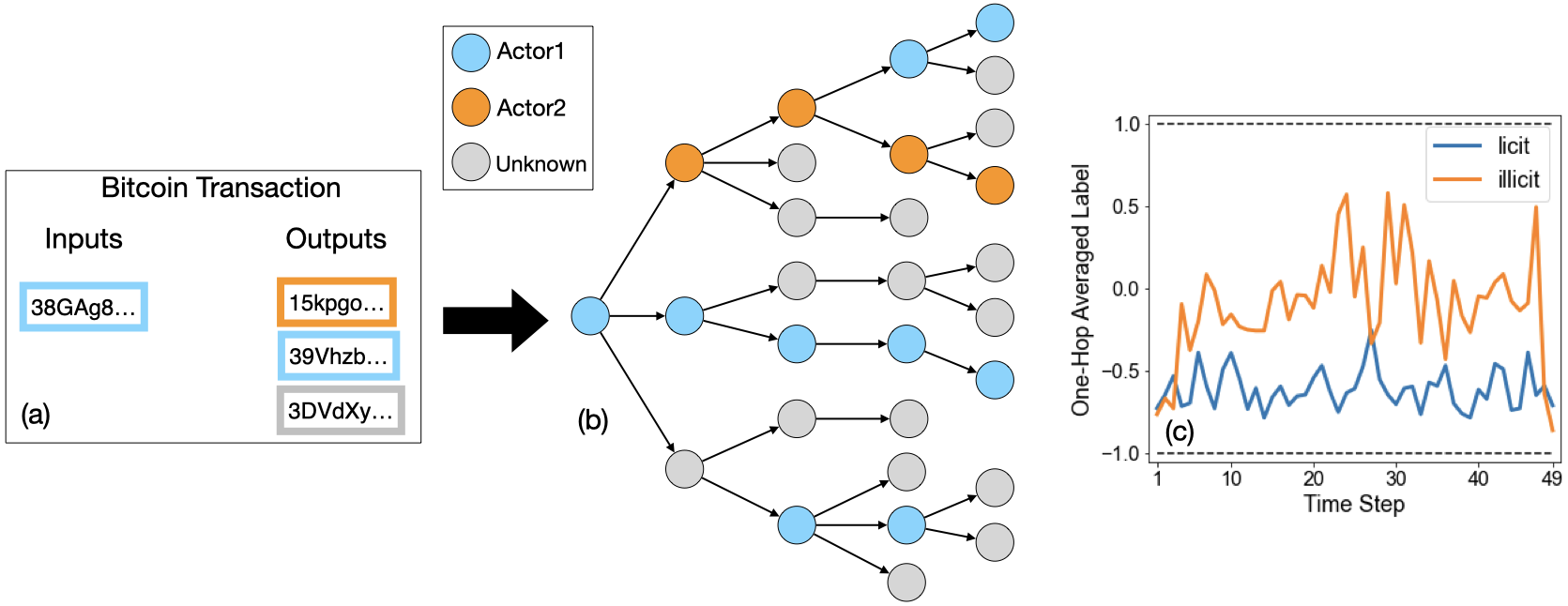}
  \caption{A Bitcoin transaction is made of inputs and outputs. Typically, among the outputs there is one that is controlled by the same entity that controls the inputs (a). This has the effect of producing chains of transactions, each of which has been created and then broadcasted to the Bitcoin network by the same entity (b). Licit and illicit nodes in the Elliptic dataset show a different behavior in the one-hop averaged labels.}
  \label{fig:propagation}
\end{figure}
\par By following the change, long chains of transactions can be built - where each transaction has been broadcasted to the Bitcoin network by a single entity. Building a chain of transactions, starting from a given labelled transaction, can be seen as propagating the label along a specific direction as shown in Figure \ref{fig:propagation}a,b. In the Elliptic dataset, the labels of the various known entities are grouped together under the labels of "licit" and "illicit" instead of being specific to an actor. However, the logic around the propagation of labels through the change address still applies. Figure \ref{fig:propagation}c, obtained after mapping "licit" $\rightarrow$ -1, "unknown" $\rightarrow$ 0, "illicit" $\rightarrow$ 1 shows that the average label at a distance of one-hop from a center node can indeed be a useful feature for the classification problem.

\subsection{Setup}
\label{sec:setup}

\begin{table}
  \centering
  \caption{Model architectures for the experiments (transductive setting).}
  \resizebox{0.7\linewidth}{!}{
  \label{tab:architeture}
  \begin{tabular}{c|cccc}
    \toprule
    Layer &  \# Neurons & Activation Function &  Parameters\\
    \midrule
    Graph Convolution  & 16 (100 Elliptic) & RELU & - \\
    Dropout & - & - & rate=0.5 \\
    Graph Convolution & 16 (100 Elliptic) & RELU & - \\
    Dropout & - & - & rate=0.5 \\
    Dense (output) & \# classes & Softmax & - \\
    \midrule
    learning rate & - & - & 0.01 \\
    early stopping - patience & - & - & 10 (30 Elliptic)\\
  \bottomrule
\end{tabular}
}
\end{table}
The Label-GCN model was trained after modifying the Stellargraph library \cite{Stellargraph2018} to include a layer implementing eq.~(\ref{e_gcnmodified1}).
\par The experiments were performed in both a transductive (CORA, PubMed, CitSeer, Elliptic) and inductive (Elliptic) setting. The models architectures for the transductive setting are shown in Table \ref{tab:architeture}. The CORA, CiteSeer and PubMed datasets were trained with two hidden layers with 16 neurons each, the Elliptic dataset was trained with two hidden layers of 100 neurons each. For each model, the number of epochs for training were chosen using early stopping with a set patience. The convolution in the first layer was either calculated using eq.~(\ref{e14}), for the standard GCN, or the modified convolution, eq.~(\ref{e_gcnmodified1}), for Label-GCN. 
\par
\begin{figure}[h]
  \centering
  \includegraphics[width=0.7\linewidth]{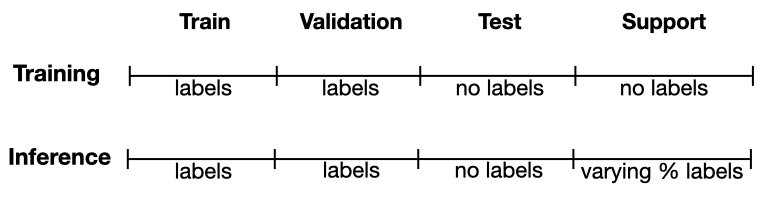}
  \caption{Summary of the setup used for evaluating the Label-GCN approach.}
  \label{fig:training_setup}
\end{figure}

 For a fair evaluation of the Label-GCN approach, the graphs were split in four different sets: a training, validation, test and support set. During the training phase of Label-GCN, the model can only see the labels associated with the training and validation set. Through these labels, the model learns how much it can rely on the neighbors' labels to make a prediction. During the inference phase, the model is also able to see a given fraction of the labels from the support set (keeping the test set always unseen). As the fraction of labels that are available through the support set increases, our results show that the model performance always improves. A summary of this "workflow" is shown in Figure \ref{fig:training_setup}. 
\par For the Elliptic dataset in the inductive setting, following \cite{Weber2019} the model was trained for 1000 epochs using a learning rate of 0.001, still using 100 neurons for the hidden layers. Finally, the label features were one-hot encoded for the CORA, CiteSeer and PubMed datasets. For the Elliptic dataset, the mapping "licit" $\rightarrow$ -1, "unknown" $\rightarrow$ 0, "illicit" $\rightarrow$ 1 was applied.

\subsection{Results}
\begin{table*}[t]
  \centering  
 \resizebox{0.85\columnwidth}{!}{  
  \begin{tabular}{c|c|c|cccc}
     &  \small{Label \% (total)} & \small{Label \% (support)} & \textbf{CORA} & \textbf{CiteSeer} &  \textbf{PubMed} \\ 
     \midrule
    \textbf{GCN} & - & - & $79.3 \pm 2.9$ & $64.8 \pm 3.4$ & $77.2 \pm 3.2$ \\
    \hline
    \multirow{6}{*}{\textbf{Label-GCN}}& 10/7/1 & 0 & $79.9 \pm 2.8$ & $64.5 \pm 3.5$ &  $77.5 \pm 3.0$\\
    & ${15}$ & 6/9/16 & $80.4 \pm 2.8$ & $65.1 \pm 3.4$ & $81.4 \pm 1.9$ \\
    & ${30}$ & 25/27/33 & $82.1 \pm 2.7$ & $66.4 \pm 3.4$ & $82.9 \pm 1.5$ \\
    & ${60}$ & 62/64/66 & $84.7 \pm 2.4$ & $68.4 \pm 3.2$ & $84.0 \pm 1.3$ \\
    & ${90}$ & 100 & $86.4 \pm 2.2$ & $70.3 \pm 3.1$ & $84.3 \pm 1.3$\\
    \bottomrule
  \end{tabular}
  }
  \caption{Mean  test  set  accuracy  and  standard  deviation for the CORA, CiteSeer and PubMed datasets. For Label-GCN, the total fraction of labels that are available during inference is also shown. Given the different splits of Table \ref{tab:1}, when the support set is fully unlabelled (0\%) the datasets are 10\%/7\%/1\% labelled for the CORA, CiteSeer and PubMed datasets respectively.}
  \label{tab:inductive}
\end{table*}

\begin{table*}[h]
  \centering  
 \resizebox{\columnwidth}{!}{
  \begin{tabular}{c|c|c|ccc|c}
   & & & \multicolumn{4}{c} {\textbf{Elliptic}} \\
   \hline
       & & & \multicolumn{3}{c|} {Illicit} & \\
       &  Label \% (total) & Label \% (support) & Precision & Recall &  F$_1$ & Accuracy\\ 
     \midrule
    \textbf{GCN} & - & - & $86.5 \pm 1.7$ & $72.8 \pm 2.2$ & $79.0 \pm 1.1$ & $96.2 \pm 0.2$\\
    \hline
    \multirow{7}{*}{\textbf{Label-GCN}}& ${20}$ & 0 & $86.3 \pm 1.4$ & $74.6 \pm 2.1$ & $80.0 \pm 1.6$ & $96.4 \pm 0.3$ \\
    & ${50}$ & 50 & $89.1 \pm 1.4$ & $74.7 \pm 2.3$ & $81.3 \pm 1.7$ & $96.6 \pm 0.3$ \\
    & ${60}$ & 67 & $89.6 \pm 1.6$ & $74.9 \pm 2.1$ & $81.6 \pm 1.6$ & $96.7 \pm 0.3$ \\
    & ${70}$ & 83 & $90.4 \pm 1.4$ & $75.2 \pm 2.5$ & $82.0 \pm 1.7$ & $96.8 \pm 0.3$ \\
    & ${80}$ & 100 & $90.7 \pm 1.5$ & $75.2 \pm 2.4$ & $82.2 \pm 1.7$ & $96.8 \pm 0.3$ \\
    \bottomrule
  \end{tabular}
  }
  \caption{Results of the experiment for the Elliptic dataset (transductive setting).}
  \label{tab:inductive_elliptic}
\end{table*}

\subsubsection{Transductive setting}
Results for the CORA, CiteSeer and PubMed datasets are summarised in Table \ref{tab:inductive}. The accuracy and standard deviation for the CORA and CiteSeer models were calculated using 100 random train/validation/test/support splits and 10 random initializations for each. The accuracy and standard deviation for the PubMed dataset were calculated using 20 splits and 10 random initializations. 
\par The results for the Elliptic dataset are summarised in Table \ref{tab:inductive_elliptic}. For this experiment, the accuracy and standard deviation was calculated using 5 random train/validation/test/support splits and 2 random initializations. The experiment was treated as an anomaly detection problem on the "illicit" class. 
\par Overall, both tables show that the performance of Label-GCN is superior to the performance of the GCN model alone, with the trend improving in favour of Label-GCN as more labels become available during the inference phase.

\subsubsection{Inductive setting}

\begin{table*}[b]
  \centering  
   \resizebox{0.95\columnwidth}{!}{
  \begin{tabular}{c|cccc|c}
    \toprule
     & \multicolumn{4}{|c|} {Illicit} & \\
    Method &  Precision &  Recall &  $F_1$ & $F_{1, t\geq t_{43}}$ & Accuracy\\
    \midrule
    Logistic Regr$^{\small{\textit{AF}}}$ & $29.5 \pm 0.0$ & $65.8 \pm 0.0$ &  $40.8 \pm 0.0$ & $5.0 \pm 0.0$ & $87.6 \pm 0.0$ \\
    Logistic Regr\small{\textit{{$^{AF+NE-GCN}$}}} & $83.8 \pm 2.7$ & $39.0 \pm 2.6$ & $53.1 \pm 2.5$ & $1.7 \pm 0.9$ & $95.5 \pm 0.2$ \\
    Logistic Regr\small{\textit{$^{AF+NE-Label-GCN}$}} & $92.9\pm 1.2$ & $57.9 \pm 3.0$ &  $71.3 \pm 2.2$ & $22.2 \pm 5.2$ & $97.0 \pm 0.2$ \\
    RandomForest\small{\textit{$^{AF}$}} & $92.7 \pm 1.3$ &  $\bf{72.1 \pm 0.2}$ &  $\bf{81.1 \pm 0.5}$ & $3.0 \pm 0.1$ & $\bf{97.8 \pm 0.1}$ \\
    RandomForest\small{\textit{$^{AF+NE-GCN}$}} & $94.1 \pm 1.7$ & $67.7 \pm 0.6$ & $78.7 \pm 0.7$ & $3.0 \pm 0.5$ & $97.6 \pm 0.1$ \\
    RandomForest\small{\textit{$^{AF+NE-Label-GCN}$}} & $\bf{97.4 \pm 1.0}$ & $61.5 \pm 2.7$ & $75.4 \pm 2.2$ & $8.3 \pm 4.0$ & $97.4 \pm 0.2$ \\
    \midrule
    GCN  & $61.6 \pm 3.8$  & $52.2 \pm 1.6$ & $56.4 \pm 1.4$ & $1.5 \pm 0.6$ &  $94.7 \pm 0.3$ \\
    Label-GCN  & $83.3 \pm 3.0$ &  $69.2 \pm 2.2$ &  $75.5 \pm 0.3$ & $\bf{36.4 \pm 3.2}$ & $97.1 \pm 0.1$ \\
  \bottomrule
\end{tabular}
}
  \caption{Results of the experiment for the Elliptic dataset (inductive setting). Following \cite{Weber2019},  $AF$ refers to all features available in the dataset. \small{\textit{NE-GCN}} denotes the node embeddings computed through GCN, and \small{\textit{NE-Label-GCN}} the node embeddings computed through Label-GCN. The bottom part of the table shows the performance of the GCN and Label-GCN models alone. The column $F_{1, t\geq t_{43}}$ shows the model performance after the dark market shutdown.}
  \label{tab:results}
\end{table*}

Label-GCN was tested in an inductive setting on the Elliptic dataset. Following \cite{Weber2019}, various models were compared against each other. In particular, Label-GCN was compared against standard GCN, logistic regression (LR) and random forest (RF). All models were trained using the first 34 time steps $t_1\dots t_{34}$ of the dataset and tested against the remaining 15 time steps $t_{35}\dots t_{49}$. Default parameters were used for LR, the RF model was run with \textit{n\_estimators}=50 and \textit{max\_features}=50 (sklearn library, version 0.23.2). In order to mitigate the imbalance in the "illicit" class, for GCN and Label-GCN the illicit nodes where over-sampled by a factor x6. 

\par The testing phase was achieved in the following manner: at each time step $t_j$ ($j > 34$), the adjacency matrix was updated to reflect the graph corresponding to all times $t <= t_j$. Then, for each test node at time $t=t_j$, full information of the node labels in the graph ($t <= t_j$) was assumed. In other words, for each test node 100\% of the labels were made available, except for the test node itself. 

Table \ref{tab:results} presents the results of the experiment. They are overall in line with \cite{Weber2019}. The performance of the standard GCN model is significantly improved by the addition of the neighbors' label as a learning feature, through Label-GCN. However, the use of node embeddings from Label-GCN only leads to a considerable improvement of the performance for LR, while no such improvement is observed for RF, in terms of F1-score. In general, it is observed that using embeddings from GCN results in a higher precision score, at the expense of recall. Interestingly, whilst all models suffer from the dark market shutdown at $t = t_{43}$ \cite{Weber2019}, label propagation aids the robustness of the models against occurrences such as this one. This is shown through the column $F_{1, t\geq t_{43}}$ of Table \ref{tab:results}, where the $F_1$-score after $t = t_{43}$ shows a measurable improvement when using Label-GCN, despite still being low in absolute values. Although feature engineering and further optimizations could lead to better results, these are outside the scope of the present paper.

\section{Conclusion}
\label{sec:conc}
We have developed Label-GCN, a modification of GCN that allows the model to learn from available labels. We have shown that this can lead to substantial improvements of the model performance, when labels are available during both the training and inference phases. Implementation of the Label-GCN model requires minor changes compared to the standard GCN implementation, without the addition of hyper-parameters, and works in both a transductive and inductive setting. We expect Label-GCN to be also applicable to the many variants of GCN that have been proposed over time, with potential benefits especially in terms of applicability to large scale applications.

\section*{Acknowledgements}
We thank useful discussions with Jonty Page.

\small

\bibliographystyle{acm}
\bibliography{library}

\begin{thebibliography}{10}

\bibitem{Androulaki2013}
{\sc Androulaki, E., Karame, G.~O., Roeschlin, M., Scherer, T., and Capkun, S.}
\newblock Evaluating user privacy in bitcoin.
\newblock In {\em International Conference on Financial Cryptography and Data
  Security\/} (2013), Springer, pp.~34--51.

\bibitem{Chen2018}
{\sc Chen, J., Ma, T., and Xiao, C.}
\newblock Fastgcn: fast learning with graph convolutional networks via
  importance sampling.
\newblock {\em arXiv preprint arXiv:1801.10247\/} (2018).

\bibitem{Stellargraph2018}
{\sc {CSIRO's Data61}}.
\newblock Stellargraph machine learning library.
\newblock \url{https://github.com/stellargraph/stellargraph}, 2018.

\bibitem{Dong2020}
{\sc Dong, H., Chen, J., Feng, F., He, X., Bi, S., Ding, Z., and Cui, P.}
\newblock On the equivalence of decoupled graph convolution network and label
  propagation.
\newblock {\em arXiv preprint arXiv:2010.12408\/} (2020).

\bibitem{Du2013}
{\sc Du, J., Zhu, F., and Lim, E.-P.}
\newblock Dynamic label propagation in social networks.
\newblock In {\em Database Systems for Advanced Applications\/} (Berlin,
  Heidelberg, 2013), W.~Meng, L.~Feng, S.~Bressan, W.~Winiwarter, and W.~Song,
  Eds., Springer Berlin Heidelberg, pp.~194--209.

\bibitem{Elliptic2019}
{\sc Elliptic}.
\newblock {Elliptic Data Set}.
\newblock \url{https://www.kaggle.com/ellipticco/elliptic-data-set}, 2019.

\bibitem{Gong2017}
{\sc Gong, C., Tao, D., Liu, W., Liu, L., and Yang, J.}
\newblock Label propagation via teaching-to-learn and learning-to-teach.
\newblock {\em IEEE transactions on neural networks and learning systems 28}, 6
  (2016), 1452--1465.

\bibitem{Hamilton2017}
{\sc Hamilton, W.~L., Ying, R., and Leskovec, J.}
\newblock Inductive representation learning on large graphs.
\newblock {\em arXiv preprint arXiv:1706.02216\/} (2017).

\bibitem{Harrigan2016}
{\sc Harrigan, M., and Fretter, C.}
\newblock The unreasonable effectiveness of address clustering.
\newblock In {\em 2016 Intl IEEE Conferences on Ubiquitous Intelligence \&
  Computing, Advanced and Trusted Computing, Scalable Computing and
  Communications, Cloud and Big Data Computing, Internet of People, and Smart
  World Congress\/} (2016), IEEE, pp.~368--373.

\bibitem{Hu2019}
{\sc Hu, Y., Seneviratne, S., Thilakarathna, K., Fukuda, K., and Seneviratne,
  A.}
\newblock Characterizing and detecting money laundering activities on the
  bitcoin network.
\newblock {\em arXiv preprint arXiv:1912.12060\/} (2019).

\bibitem{Szummer2002}
{\sc Jaakkola, M. S.~T., and Szummer, M.}
\newblock Partially labeled classification with markov random walks.
\newblock {\em Advances in neural information processing systems (NIPS) 14\/}
  (2002), 945--952.

\bibitem{Karasuyama2013}
{\sc Karasuyama, M., and Mamitsuka, H.}
\newblock Manifold-based similarity adaptation for label propagation.
\newblock {\em Advances in neural information processing systems 26\/} (2013),
  1547--1555.

\bibitem{Kipf2017}
{\sc Kipf, T.~N., and Welling, M.}
\newblock Semi-supervised classification with graph convolutional networks.
\newblock {\em arXiv preprint arXiv:1609.02907\/} (2016).

\bibitem{Meiklejohn2016}
{\sc Meiklejohn, S., Pomarole, M., Jordan, G., Levchenko, K., McCoy, D.,
  Voelker, G.~M., and Savage, S.}
\newblock A fistful of bitcoins: characterizing payments among men with no
  names.
\newblock In {\em Proceedings of the 2013 conference on Internet measurement
  conference\/} (2013), pp.~127--140.

\bibitem{Namata2012}
{\sc Namata, G., London, B., Getoor, L., Huang, B., and EDU, U.}
\newblock Query-driven active surveying for collective classification.
\newblock In {\em 10th International Workshop on Mining and Learning with
  Graphs\/} (2012), vol.~8.

\bibitem{Oliveira2021}
{\sc Oliveira, C., Torres, J., Silva, M.~I., Apar{\'\i}cio, D., Ascens{\~a}o,
  J.~T., and Bizarro, P.}
\newblock Guiltywalker: Distance to illicit nodes in the bitcoin network.
\newblock {\em arXiv preprint arXiv:2102.05373\/} (2021).

\bibitem{Ragesh2020}
{\sc Ragesh, R., Sellamanickam, S., Lingam, V., and Iyer, A.}
\newblock A graph convolutional network composition framework for
  semi-supervised classification.
\newblock {\em arXiv preprint arXiv:2004.03994\/} (2020).

\bibitem{Raghavan2007}
{\sc Raghavan, U.~N., Albert, R., and Kumara, S.}
\newblock Near linear time algorithm to detect community structures in
  large-scale networks.
\newblock {\em Physical review E 76}, 3 (2007), 036106.

\bibitem{Reid2013}
{\sc Reid, F., and Harrigan, M.}
\newblock An analysis of anonymity in the bitcoin system.
\newblock In {\em Security and privacy in social networks}. Springer, 2013,
  pp.~197--223.

\bibitem{Ron2013}
{\sc Ron, D., and Shamir, A.}
\newblock Quantitative analysis of the full bitcoin transaction graph.
\newblock In {\em International Conference on Financial Cryptography and Data
  Security\/} (2013), Springer, pp.~6--24.

\bibitem{Sen2008}
{\sc Sen, P., Namata, G., Bilgic, M., Getoor, L., Galligher, B., and
  Eliassi-Rad, T.}
\newblock Collective classification in network data.
\newblock {\em AI magazine 29}, 3 (2008), 93--93.

\bibitem{Shchur2019}
{\sc Shchur, O., Mumme, M., Bojchevski, A., and G{\"u}nnemann, S.}
\newblock Pitfalls of graph neural network evaluation.
\newblock {\em arXiv preprint arXiv:1811.05868\/} (2018).

\bibitem{Viswanathan2019}
{\sc Viswanathan, K., Sachdeva, S., Tomkins, A., and Ravi, S.}
\newblock Improved semi-supervised learning with multiple graphs.
\newblock In {\em The 22nd International Conference on Artificial Intelligence
  and Statistics\/} (2019), PMLR, pp.~3032--3041.

\bibitem{Wang2013}
{\sc Wang, B., Tu, Z., and Tsotsos, J.~K.}
\newblock Dynamic label propagation for semi-supervised multi-class multi-label
  classification.
\newblock In {\em Proceedings of the IEEE international conference on computer
  vision\/} (2013), pp.~425--432.

\bibitem{Wang2020}
{\sc Wang, H., and Leskovec, J.}
\newblock Unifying graph convolutional neural networks and label propagation.
\newblock {\em arXiv preprint arXiv:2002.06755\/} (2020).

\bibitem{Wang2009}
{\sc Wang, J., Wang, F., Zhang, C., Shen, H.~C., and Quan, L.}
\newblock Linear neighborhood propagation and its applications.
\newblock {\em IEEE transactions on pattern analysis and machine intelligence
  31}, 9 (2008), 1600--1615.

\bibitem{Weber2019}
{\sc Weber, M., Domeniconi, G., Chen, J., Weidele, D. K.~I., Bellei, C.,
  Robinson, T., and Leiserson, C.~E.}
\newblock Anti-money laundering in bitcoin: Experimenting with graph
  convolutional networks for financial forensics.
\newblock {\em arXiv preprint arXiv:1908.02591\/} (2019).

\bibitem{Wu2019}
{\sc Wu, F., Souza, A., Zhang, T., Fifty, C., Yu, T., and Weinberger, K.}
\newblock Simplifying graph convolutional networks.
\newblock In {\em International conference on machine learning\/} (2019), PMLR,
  pp.~6861--6871.

\bibitem{Wu2012}
{\sc Wu, X.-M., Li, Z., So, A. M.-C., Wright, J., and Chang, S.-F.}
\newblock Learning with partially absorbing random walks.
\newblock In {\em NIPS\/} (2012), vol.~25, pp.~3077--3085.

\bibitem{Zhang2020}
{\sc Zhang, Z., Cui, P., and Zhu, W.}
\newblock Deep learning on graphs: A survey.
\newblock {\em IEEE Transactions on Knowledge and Data Engineering\/} (2020).

\bibitem{Zhou2003}
{\sc Zhou, D., Bousquet, O., Lal, T.~N., Weston, J., and Sch{\"o}lkopf, B.}
\newblock Learning with local and global consistency.
\newblock {\em Advances in neural information processing systems 16}, 16
  (2004), 321--328.

\bibitem{Zhou2018}
{\sc Zhou, J., Cui, G., Zhang, Z., Yang, C., Liu, Z., Wang, L., Li, C., and
  Sun, M.}
\newblock Graph neural networks: A review of methods and applications.
\newblock {\em arXiv preprint arXiv:1812.08434\/} (2018).

\bibitem{Zhu2002}
{\sc Zhu, X., and Ghahramani, Z.}
\newblock Learning from labeled and unlabeled data with label propagation.

\end{thebibliography}

\end{document}